\documentclass[letterpaper, 10 pt, journal, twoside]{IEEEtran}

\IEEEoverridecommandlockouts                              
                                                          
\usepackage{amsmath}
\usepackage{amsfonts}
\usepackage{amssymb}
\usepackage{balance}
\usepackage{graphicx}
\usepackage[skip=2pt,font=small]{caption}
\usepackage{subcaption}
\usepackage{siunitx}
\usepackage{color}
\usepackage{xcolor}
\usepackage{booktabs}
\usepackage{dblfloatfix}
\usepackage{url}
\usepackage{nth}
\usepackage{algorithmic}    
\usepackage[linesnumbered,algoruled,boxed,lined]{algorithm2e}
\usepackage{booktabs}
\usepackage{float}
\usepackage{caption}
\usepackage{dsfont}
\usepackage{multirow}
\usepackage{hyperref}

\usepackage{nicefrac}

\usepackage{pgfplots}
\pgfplotsset{compat=newest}
\usetikzlibrary{plotmarks}
\usetikzlibrary{babel}
\usetikzlibrary{arrows.meta}
\usepgfplotslibrary{patchplots}
\usepackage{grffile}
\usepackage{amsmath}

\usepgfplotslibrary{groupplots}
\usepgfplotslibrary{dateplot}

\usepackage{capt-of}

\usepackage{color}

\newcommand{\revision}[1]{{\color{black} #1}}
\newcommand{\rebuttal}[1]{{\color{black} #1}}

\begin{document}
%
\title{AutoTune: Controller Tuning for High-Speed Flight}
%
%
%


\author{Antonio Loquercio*, Alessandro Saviolo*, and Davide Scaramuzza%
\thanks{Manuscript received: September, 9th, 2021; Revised December, 9th, 2021;
Accepted January, 11th, 2022. This paper was recommended for publication by Editor Tamim Asfour upon evaluation of the Associate Editor and Reviewers' comments. The first two authors contributed equally. The work was done at the Robotics and Perception Group, University of Zurich, Switzerland (\url{http://rpg.ifi.uzh.ch}) and was supported by the National Centre of Competence in Research (NCCR) Robotics, through the Swiss National Science Foundation (SNSF), and the European Research Council (ERC) under the European Union’s  Horizon 2020 research and innovation programme (Grant agreement No. 864042).}%
}

\renewcommand\thefigure{\arabic{figure}}    

\makeatletter
\g@addto@macro\@maketitle{
\setcounter{figure}{0}
\centering
    \includegraphics[width=\linewidth]{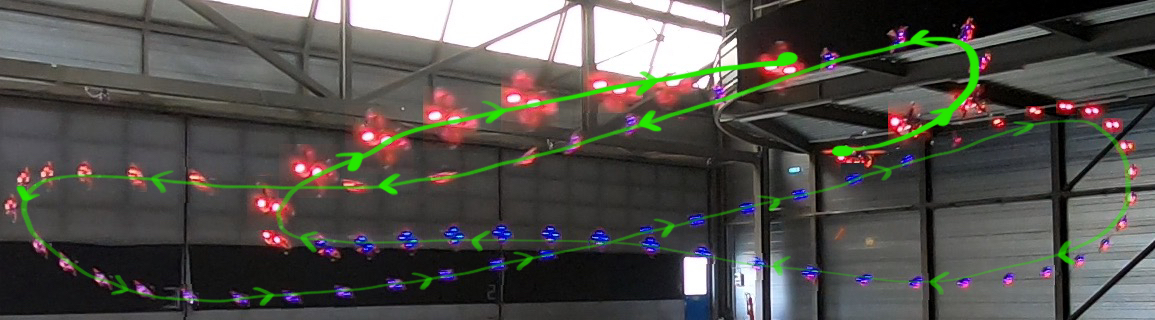}
	\captionof{figure}{Our quadrotor flies a time-optimal trajectory with speeds over \SI{50}{\kilo\meter\per\hour}.  We automatically find a controller configuration that can fly such a high-speed maneuver with a novel sampling-based method called AutoTune.  To get a better sense of the speed achieved by the quadrotor, please watch the supplementary movie\label{fig:rw_tracking}.}
	\vspace{-3ex}
}
\makeatother

\markboth{IEEE Robotics and Automation Letters. Preprint Version. Accepted
January, 2022}
{Loquercio \MakeLowercase{\textit{et al.}}: AutoTune: Controller Tuning for High-Speed Flight}

\maketitle


\begin{abstract}
Due to noisy actuation and external disturbances, tuning controllers for high-speed flight is very challenging.
In this paper, we ask the following questions:  How sensitive are controllers to tuning when tracking high-speed maneuvers? What algorithms can we use to automatically tune them?
To answer the first question, we study the relationship between parameters and performance and find out that the faster the maneuver, the more sensitive a controller becomes to its parameters.
To answer the second question, we review existing methods for controller tuning and discover that prior works often perform poorly on the task of high-speed flight. 
Therefore, we propose AutoTune, a sampling-based tuning algorithm specifically tailored to high-speed flight.
In contrast to previous work, our algorithm does not assume any prior knowledge of the drone or its optimization function and can deal with the multi-modal characteristics of the parameters' optimization space.
We thoroughly evaluate AutoTune both in simulation and in the physical world.  
In our experiments, we outperform existing tuning algorithms by up to 90\% in trajectory completion. 
The resulting controllers are tested in the AirSim Game of Drones competition, where we outperform the winner by up to 25\% in lap-time.
Finally, we validate AutoTune in real-world flights in one of the world’s largest motion-capture systems.
In these experiments, we outperform human experts on the task of parameter tuning for trajectory tracking, achieving flight speeds over \SI{50}{\kilo\meter\per\hour}
\end{abstract}
\vspace{-4ex}

\section*{Supplementary Material}

Video and code are at \href{https://youtu.be/m2q_y7C01So}{\rebuttal{https://youtu.be/m2q\_y7C01So}} and \href{https://github.com/uzh-rpg/mh_autotune}{https://github.com/uzh-rpg/mh\_autotune}.


\section{Introduction}

\begin{figure}[t]
    \centering
    \includegraphics[width=0.78\linewidth]{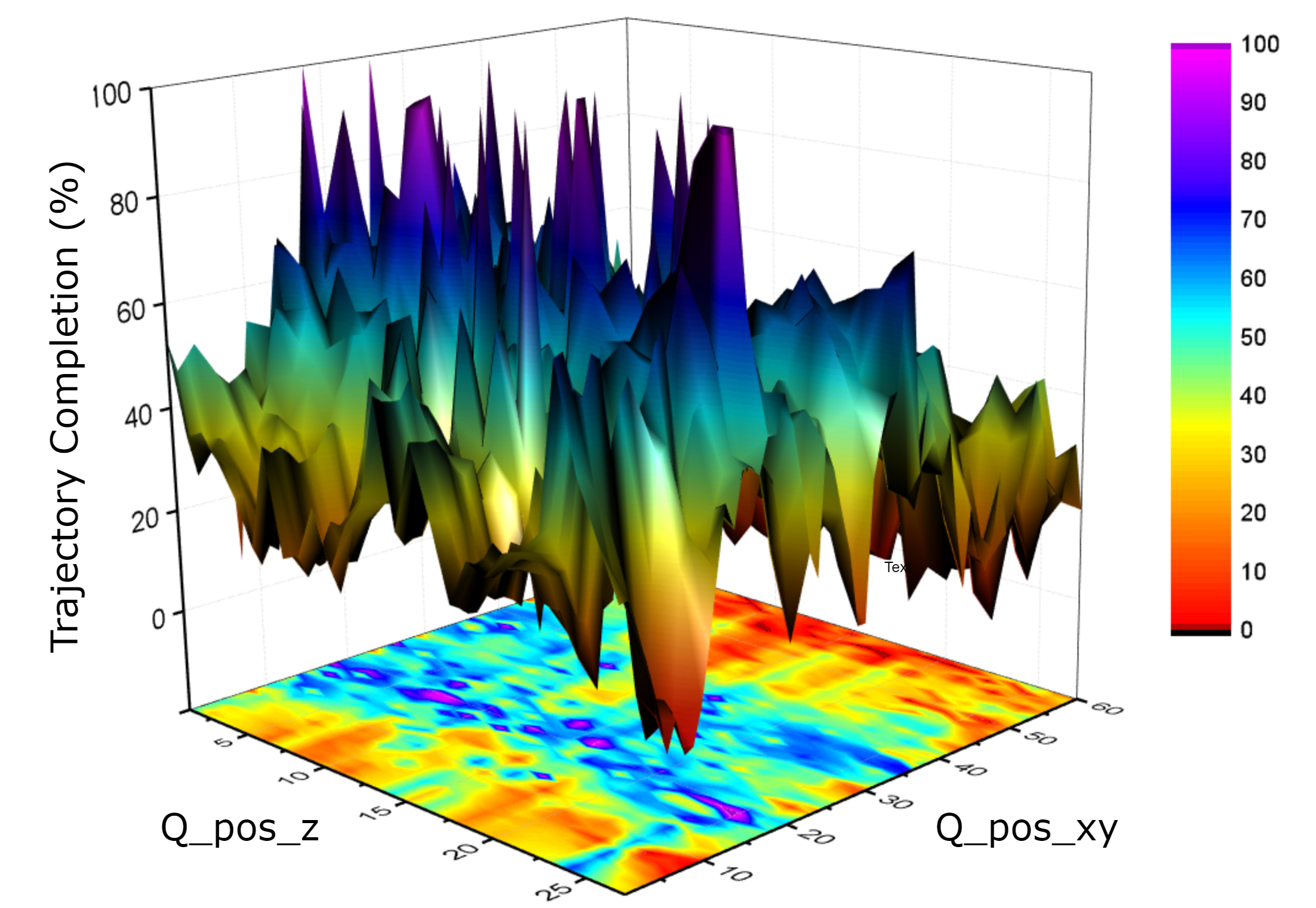}
    \caption{Trajectory completion (\%) as a function of two parameters of a model-predictive controller. \revision{The trajectory completion measures the percentage of trajectory successfully tracked by the controller.}
    The high speed and high angular accelerations required by time-optimal trajectories make the controller extremely sensitive to its parameters.}
    \vspace{-3ex}
    \label{fig:opt_landscape}
\end{figure}

Flying high-speed trajectories with a quadrotor requires the platform's controller to be meticulously tuned~\cite{Foehn2020,romero2021model}.
%
%
The complex relationship between parameters and performance, empirically shown in Fig.~\ref{fig:opt_landscape}, is caused by unavoidable factors such as imperfect modeling and external disturbances.
This work is motivated by the following questions: What are the characteristics of this optimization space? How can we automatically find controller parameters for high-speed flight?

Tuning controller parameters to fly high-speed maneuvers is difficult due to three main challenges: (i) the objective function (i.e. the relationship between controller parameters and performance) is highly non-convex (See Fig.~\ref{fig:opt_landscape}); (ii) the tuning process only relies on noisy evaluations\footnote{Due to noise the same controller parameters can yield different performance on multiple runs.} of the objective function at adaptively chosen parameters, but not to the function itself or its gradients; (iii) different parts of the trajectory, e.g. a sharp turn or a straight-line acceleration, generally require different controller behaviors, hence dynamically changing parameters.

%
The traditional approach for automatic tuning and adaptive control, generally known as MIT rule~\cite{Astrom83Autom}, requires to express the desired performance metric, \emph{e.g.} the average tracking error over the entire maneuver, as a quadratic function of controller parameters, and then optimizes the controller with gradient-based optimization~\cite{lqr_tuning_1, gradient_tuning_2, Basri15TIMC}.
However, expressing the long-term performance on a high-speed maneuver with respect to the parameters of a receding horizon controller (\emph{i.e.} the optimization function depicted in Fig.~\ref{fig:opt_landscape}) is generally intractable.
Indeed, it requires to know \emph{a priori} the exact model of the quadrotor and the disturbances acting on it during flight, \emph{e.g.} noisy actuation and aerodynamic effects.
Instead of analytically computing it, another line of work proposes to iteratively estimate the optimization function, and use the estimate to find optimal parameters~\cite{menner2020maximum, berkenkamp2016safe, marco_ICRA_2016}.
%
%
However, these methods make over-simplifying assumptions on the objective function, e.g. convexity or relative Gaussianity between observations.
Such assumptions are generally not suited for controller tuning to high-speed flight, where the function is highly non-convex (c.f. Fig.~\ref{fig:opt_landscape}).
To remove any assumption, model-free methods propose to directly search for optimal parameters using sampling.
Such methods are however built on heuristics not necessarily suited to high-speed flight and generally require thousands of iterations to converge~\cite{ga_tuning}.


In this paper, we propose a novel sampling-based algorithm specifically tailored to the problem of high-speed flight, rooted in statistical theory: AutoTune.
Given an initial, low-performance controller, AutoTune optimizes its parameters to maximize a user-defined metric, e.g. track completion.
In contrast to traditional adaptive control, \emph{e.g.} the MIT rule, it does not require to analytically express the optimization function with respect to the controller parameters,  nor assumptions about the optimization function.
Similarly to model-free sampling-based methods, AutoTune does neither require prior knowledge of the platform model and external disturbances.
However, to make sampling computationally tractable, our approach uses Metropolis-Hastings sampling (M-H)~\cite{Metropolis1953} and several strategies specifically tailored to the problem of high-speed flight.
%
%
Specifically, motivated by the observation that different parts of a trajectory require different controller behaviors, we propose a strategy to break down a trajectory into components with different behaviors, e.g. sharp descent or planar acceleration.
Despite controller parameters being different for each component, they are all optimized jointly to favor optimality over the entire trajectory.
In addition, to speed up convergence, we train a regressor to predict good initialization parameters.

\revision{We perform an extensive evaluation in two simulators and in the physical world in a large tracking arena of $30\times30\times8\SI{}{\meter}$ volume}.
In these experiments, we find out that: (i) the faster a maneuver is, the more sensitive a controller becomes to its parameters, and (ii) the optimization function is multi-modal, \emph{i.e.} multiple controller configurations lead to the desired performance. %
We empirically show that our approach can tune controllers up to 90 percentage points better than previous work in terms of trajectory completion.
We then validate the controller parameters found by AutoTune in simulation on a physical platform.
\revision{
These parameters decrease the tracking error with respect to the ones tuned by a human expert, enabling the quadrotor to achieve speeds over \SI{50}{\kilo\meter\per\hour}}.
Overall, our work makes the following contributions:
\begin{itemize}
    \item We present a novel sampling-based method for tuning quadrotor controllers on the task of high-speed flight.
    \item \revision{We show that our method outperforms existing methods for automatic controller tuning and enables quadrotors to fly time-optimal trajectories both in simulation and in the physical world in one of the world’s largest motion-capture systems}.
    \item We provide interesting insights into the relationship between the parameters of a receding horizon controller and its flight performance on high-speed maneuvers.
\end{itemize}

\section{Related work}

The simplest option available to robotic researchers for controller tuning is to use domain knowledge, i.e. experience, to tune controllers' parameters.
However, tuning by hand often translates in a tedious and time-consuming trial-and-error process, difficult even for the simplest maneuvers.
Besides, human intuition often provides an inherent bias to the experiments, which results in sub-optimal performance and calls for a more principled parameter tuning approach.

In line with adaptive control, the classic approach for controller tuning analytically finds the relationship between a performance metric, \emph{e.g.} tracking error or trajectory completion, and optimizes the parameters with gradient-based optimization~\cite{Astrom83Autom,lqr_tuning_1,gradient_tuning_2,Basri15TIMC}.
However, doing so requires to analytically derive the performance of a receding-horizon controller over a possibly long maneuver, which is intractable given the model errors, the noisy actuation, and other perturbations possibly acting on the platform during flight, \emph{e.g.} aerodynamics effects.
Approximating these effects numerically is possible for short maneuvers~\cite{lupashin2010simple,Julka15IES,CEDRO2019156}, but the more complex the maneuver or the system is, the more difficult the identification becomes, making these methods impractical for tuning controllers to fly time-optimal maneuvers.
\rebuttal{
When a precise model of the platform is not available, methods like L1 adaptive control can estimate model errors online and account for them in a reactive fashion~\cite{hanover2021performance}. 
However, these methods trade off robustness with performance, that leads to sub-optimal behaviour during high-speed motion. 
In addition, they also have hyper-parameters to tune, so they could be complementary to our approach.
}


Motivated by this difficulty, another family of approaches estimates the relationship between the controller's performance and its parameters directly from data~\cite{lqr_inverted_pendulum, berkenkamp2016safe, menner2020maximum}.
%
%
Through multiple experiments, both the estimate and the parameters are iteratively updated.
However, doing so requires making additional assumptions on the shape of the function.
The assumptions commonly used in the literature are: (i) relative normality between all observations according to some pre-defined kernel, as in Bayesian Optimization~\cite{berkenkamp2016safe, lqr_inverted_pendulum, marco_ICRA_2016}, and (ii) the function can be described by a parametric distribution, e.g. a Gaussian, as typical in inverse optimal control~\cite{menner2020maximum}.
When the relationship between parameters and performance is very complex, as it is the case for time-optimal trajectories, these assumptions generally cause a poor fitting of the function, which results in sub-optimal tuning performance.
If demonstrations by a human expert are available, another option consists of using inverse reinforcement learning~\cite{DeepakInverseRl,xiao2020appld}, but this is generally not the case with time-optimal trajectories, which can be faster than the trajectories flown by the best human pilots~\cite{Foehn2020}. 

\begin{figure*}[t]
    \centering
    \includegraphics[width=\textwidth]{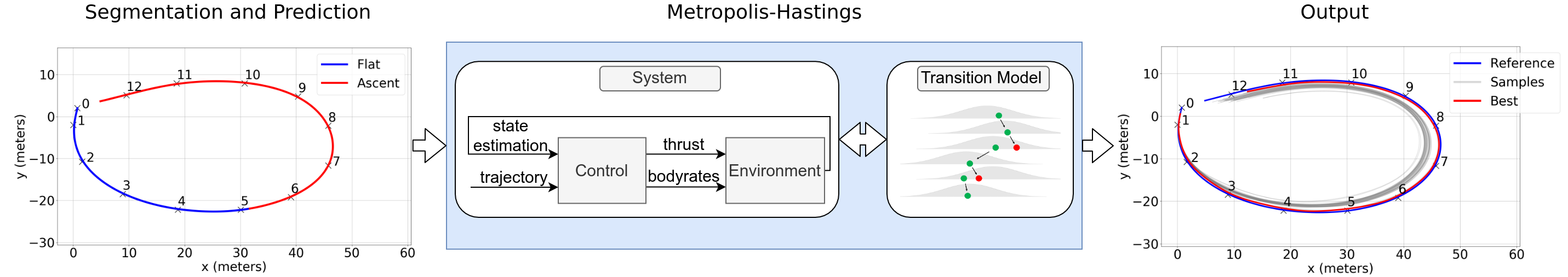}
    \caption{We compute a minimum-time trajectory passing through all waypoints~\cite{Foehn2020}. The trajectory is then segmented in parts that require different controller behaviors, and initial parameters for each segment are predicted with a regressor. The parameters are then jointly optimized with M-H sampling over multiple rollouts.} 
    \label{fig:autotune}
    \vspace{-3ex}
\end{figure*}

To relax the assumptions required by the previous methods, another family of algorithms proposes to directly search for the optimal controller parameters by sampling.
The main advantage of these algorithms is that they can deal with highly non-convex functions, like the one between parameters and performance in time-optimal trajectories.
\revision{However, while this approach has encountered high successes with grounded robots~\cite{Wang_2021_apple}, it's validity for agile systems like quadrotors is still unclear. Indeed, it was applied only on very simple tasks, \emph{e.g.} hovering~\cite{Khuwaya_ga_tuning_quadrotors}}.
The major issue is how to define the parameters sampling procedure.
\rebuttal{
We propose to do so with Metropolis-Hastings sampling~\cite{Metropolis1953}. 
Prior work successfully deployed this technique for collision-free planning in complex environments~\cite{loquercio2021learning}.
However, their approach performed a single large optimization, which is challenging and computationally expensive.
In contrast, we design a trajectory segmentation procedure to divide the optimization in small sub-problems, which are faster and easier to solve.}


\section{Preliminaries and Overview}

We define the task of high-speed flight through a series of waypoints as finding a policy minimizing the following cost:
\begin{align} \label{eq:problem_formulation}
    \min_\pi J(\pi) &=
    \mathbb{E}_{\rho(\pi)}
    \left[ t(\pi) \right], \\
    &\text{subject to} \ \|\textbf{u}[k]\| \leq \textbf{u}_{c} \label{eq:actuation_constr}\\
                       &\qquad \qquad  \textbf{s}[k+1] = f(\textbf{s}[k],\textbf{u}[k]),  \label{eq:dynamics}
\end{align}
where $\textbf{s}[k]$ is the quadrotor's state at time $k$, $\textbf{u}[k]$ is the input, $\rho(\pi)$ is the distribution of possible trajectories induced by the controller $\pi$, and $t(\pi)$ is the time required to fly through all waypoints.
The solution of Eq.~\eqref{eq:problem_formulation} is a policy that minimizes the time to pass through all waypoints by respecting the platform dynamics (Eq.~\eqref{eq:dynamics}) and actuation constraints (Eq.~\eqref{eq:actuation_constr}).
Given the series of waypoints and the platform's specifics, we approximate a solution to Eq.~\eqref{eq:problem_formulation} with non-convex optimization~\cite{Foehn2020}, and then track the resulting trajectory $\mbox{\boldmath$\tau$}_r$ with a receding-horizon controller.

Analytical controllers, e.g. the Linear Quadratic Regulator (LQR), aim to solve the tracking problem directly.
\revision{However, they disregard the platform dynamics and actuation constraints, resulting in poor controller performance when there is a mismatch between the model and the physical system}~\cite{borrelli2017predictive}.
\revision{In these conditions, the controller gains are tuned to maximize performance on the physical system, generally with iterative learning approaches~\cite{lupashin2010simple}.}
Conversely, model predictive controllers (MPC) propose to solve a finite horizon version of the tracking problem in a receding horizon fashion:
\begin{align} \label{eq:problem_formulation_mpc}
    \pi (\mathbf{x}[t_0])  &= \min_U
    \left[
    \sum_{k=t_0}^{t_0 + t_h} \textbf{x}[k]^\top \mbox{\boldmath$Q$} \textbf{x}[k] +
    \textbf{u}[k]^\top \mbox{\boldmath$R$} \textbf{u}[k]
    \right] \\
    &\text{subject to} \ \|\textbf{u}[k]\| \leq \textbf{u}_{c} \notag\\
                       &\qquad \qquad  \textbf{x}[k+1] = f(\textbf{x}[k],\textbf{u}[k]),  \notag
\end{align}
where $x[k] = \mbox{\boldmath$\tau$}_r[k] - \textbf{s}[k]$ denotes the difference between the state of the platform and the corresponding reference at time $k$, $\mbox{\boldmath$Q$}$ and $\mbox{\boldmath$R$}$ are the state and input cost matrices, and $t_h$ is the horizon length, generally much smaller than the entire trajectory duration $t$.
In contrast to the LQR, the platform constraints and dynamics are directly taken into account by the controller.
\revision{This results in better behavior in case of mismatch between the model and the physical system}~\cite{borrelli2017predictive}.
This approach, however, requires to tune the controller parameters $\mbox{\boldmath$Q$}$, $\mbox{\boldmath$R$}$, and $t_h$ to minimize a user-defined metric, \emph{e.g.} the tracking error, over the entire trajectory~\footnote{Parameters are equivalent up to scale. To account for this effect, we keep the cost on inputs $\mbox{\boldmath$R$}$ constant}.
Tuning these parameters is challenging since it is not possible to analytically find the relationship between them and the long-horizon cost, as possible for LQR controllers via the MIT rule~\cite{Astrom83Autom}.
%

\rebuttal{Other approaches for high-speed trajectory tracking are non-linear geometric controllers and adaptive ones. However, a recent study has shown that in the absence of a precise model of the system, model-predictive control generally outperforms other approaches in the task of high-speed flight~\cite{sun2021comparative}.}

%
%
To improve the performance of MPC, we propose a strategy to split a trajectory into parts that require different controller behavior, hence different parameters. 
The parameters are optimized jointly to favor global optimization.
Additionally, we initialize the search from a good guess of parameters to reduce sampling time.
These parameters are predicted by a regressor trained on previously optimized tracks.
Fig.~\ref{fig:autotune} shows a summary of the proposed approach to tune controllers for high-speed flight.
\rebuttal{Despite being specific for MPC, we hypothesize that similar conclusions could be drawn when tuning different controllers.}
The next section presents each aspect of our method in detail.

\section{Method}

\subsection{Metropolis-Hastings Sampling}\label{sec:mh_sampling}

In statistics, the Metropolis-Hastings (M-H) algorithm~\cite{hastings70} is used to obtain a sequence of random samples from a desired distribution $P(w)$ which can't be directly accessed.
To generate the samples, the M-H algorithm requires a score function $d(w)$ which is proportional to the density $P(w)$.
Samples are produced in an iterative fashion: the next sample $w_{t+1}$ comes from a distribution $t(w_{t+1} | w_t)$, referred to as transition model, which only depends on the current sample $w_t$.
\rebuttal{As transition model $t(w_{t+1} | w_t)$ we select a Gaussian with constant variance $\sigma=5$ centered on $w_t$. We keep this transition model fixed for all experiments}. 
The next sample $w_{t+1}$ is then accepted and used for the next iteration, or it is rejected, discarded, and the current sample $w_t$ is re-used for the next iteration.
Specifically, the sample is accepted with probability equal to 
\begin{equation}
    \alpha = min(1, \frac{d(w_{t+1})}{d(w_t)}) = min(1, \frac{P(w_{t+1})}{P(w_t)}).
\end{equation}
Therefore, M-H always accepts a sample with a higher score.
However, the move to a sample with a smaller score will sometimes be rejected, and the higher the drop in score $\frac{1}{\alpha}$, the smaller the probability of acceptance.
Therefore, many samples come from the high-density regions of $P(w)$, while relatively few from the low-density regions.
Intuitively, this is why the empirical sample distribution $\hat{P}(w)$ approximates the target distribution $P(w)$.

In this work, we use the M-H algorithm to find the parameters of a controller flying time-optimal trajectories.
\rebuttal{In this case, $P(w) = 1/Z d(w)$, where $w$ are MPC parameters, $Z$ is an unknown normalization factor, and $d(w)$ is the score function:
\begin{equation}\label{eq:score_fun}
    d(w) = \exp (-m(w)),
\end{equation}
where $m(w)$ is a metric measuring the performance (\emph{e.g.} time) the controller accumulates over the entire trajectory.
According to $P(w)$, the points with maximum probability are the ones with higher score.
}
However, in the task of controller tuning, we are not interested in approximating the distribution of controller parameters $\hat{P}(w)$, but to find, with as few samples as possible, parameters that enable tracking a trajectory accurately.
\rebuttal{We continue the sampling procedure up to when we find a solution satisfying some user-defined performance metrics, \emph{e.g.} tracking error or trajectory completion. When found, we re-evaluate the solution for four times to account for the randomness of the simulation.}

\rebuttal{This setup makes the use of Metropolis-Hasting sampling equivalent to simulated annealing with constant temperature~\cite{Kirkpatrick83Science}.
Despite varying-temperature simulated annealing providing the asymptotic guarantee of global optimality, it generally requires a significantly larger number of samples with respect to its constant-temperature counterpart~\cite{Fielding00Opt} and a specifically designed heuristic to define the cooling function~\cite{Kirkpatrick83Science}.
Therefore, since we are not interested in the global optimum but only in a controller configuration satisfying a user-defined performance metric, we keep the temperature to a constant value.
In addition, due to the temperature-based sampling, simulated annealing does not provide the possibility to approximate the distribution of controller parameters (c.f. Fig.~\ref{fig:opt_landscape}). Conversely, MH can approximate the distribution and used to study the characteristics of the optimization space.}

\subsection{Scoring Performance with Time}

According to Eq.~\eqref{eq:problem_formulation}, we define the metric $m(w)$ in Eq.~\eqref{eq:score_fun} to be the time $t$ to pass all waypoints, \emph{i.e.} $m(w) = J(\pi(w))$.
However, it is not clear how to define this metric when the drone misses a waypoint or crashes before the end of the trajectory.
To solve this problem, we propose to stop the experiment whenever the drone misses a waypoint or crashes.
In this case, a penalty equal to the shortest path between the drone position and all further waypoints is added to the time.
In such a way, it is possible to distinguish between parameters $w$ that make the drone crash in the early stage of a trajectory and the ones that can complete the trajectory to the end.
%

\begin{table*}[t]
    \centering
    \begin{tabular}{c c c c c c c c c c c c }
        \toprule
        \multirow{2}{*}{Track} & \multirow{2}{*}{Max Vel} & \multirow{2}{*}{Random} & \multicolumn{5}{c}{Bayesian Optimization} & \multirow{2}{*}{\rebuttal{PSO}} & \multirow{2}{*}{\rebuttal{CMA-ES}} & \multirow{2}{*}{\rebuttal{Gradient}} & \multirow{2}{*}{AutoTune}\\
        \cmidrule(lr){4-8}
        & [$\SI{}{\meter\per\second}$] & Search & Mat & LocalPer & RQ & SQ & Per & \rebuttal{(1000)} & \rebuttal{(1000)} & \rebuttal{Boosting} & (Ours) \\
        \midrule
        Circle & $20$  & 100 & 100 & 100 & 100 & 100 & 100 & \rebuttal{100} & \rebuttal{100} & \rebuttal{100} & \textbf{100}\\
        Circle & $34$ & 65 & 80 & 65 & 60 & 60 & 30 & \rebuttal{100} & \rebuttal{100} & \rebuttal{65} & \textbf{100}\\
        Drop & $20$  & 40 & 100 & 50 & 50 & 50 & 40 & \rebuttal{100} & \rebuttal{100} & \rebuttal{50} & \textbf{100}\\
        Flip & $16$  & 80 & 80 & 80 & 80 & 80 & 80 & \rebuttal{100} & \rebuttal{100} & \rebuttal{80} & \textbf{100}\\
        Spiral & $53$  & 0 & 10 & 0 & 0 & 10 & 0 & \rebuttal{10} & \rebuttal{10} & \rebuttal{10} & \textbf{100}\\
        Qualifier & $20$  & 50 & 75 & 60 & 60 & 60 & 30 & \rebuttal{75} & \rebuttal{100} & \rebuttal{65} & \textbf{100}\\
        Final & $22$ & 10 & 40 & 40 & 30 & 30 & 25 & \rebuttal{40} & \rebuttal{100} & \rebuttal{40} & \textbf{100}\\
        \bottomrule
    \end{tabular}
    \vspace{1ex}
    \caption{Comparison of AutoTune with the baselines. All approaches have a maximum budget of $200$ \rebuttal{samples}, except PSO and CMA-ES, whose budget is 1000 samples. While all baselines perform well on easy maneuvers, their performance drops when the speed and angular acceleration required by the trajectory increases. Conversely, AutoTune can always find parameters to fly the entire trajectory.}
    \vspace{-2ex}
    \label{tab:optimization_results}
\end{table*}

\subsection{Trajectory Segmentation} \label{subsec:segmentation}

\begin{figure}[t]
    \includegraphics[width=0.9\linewidth]{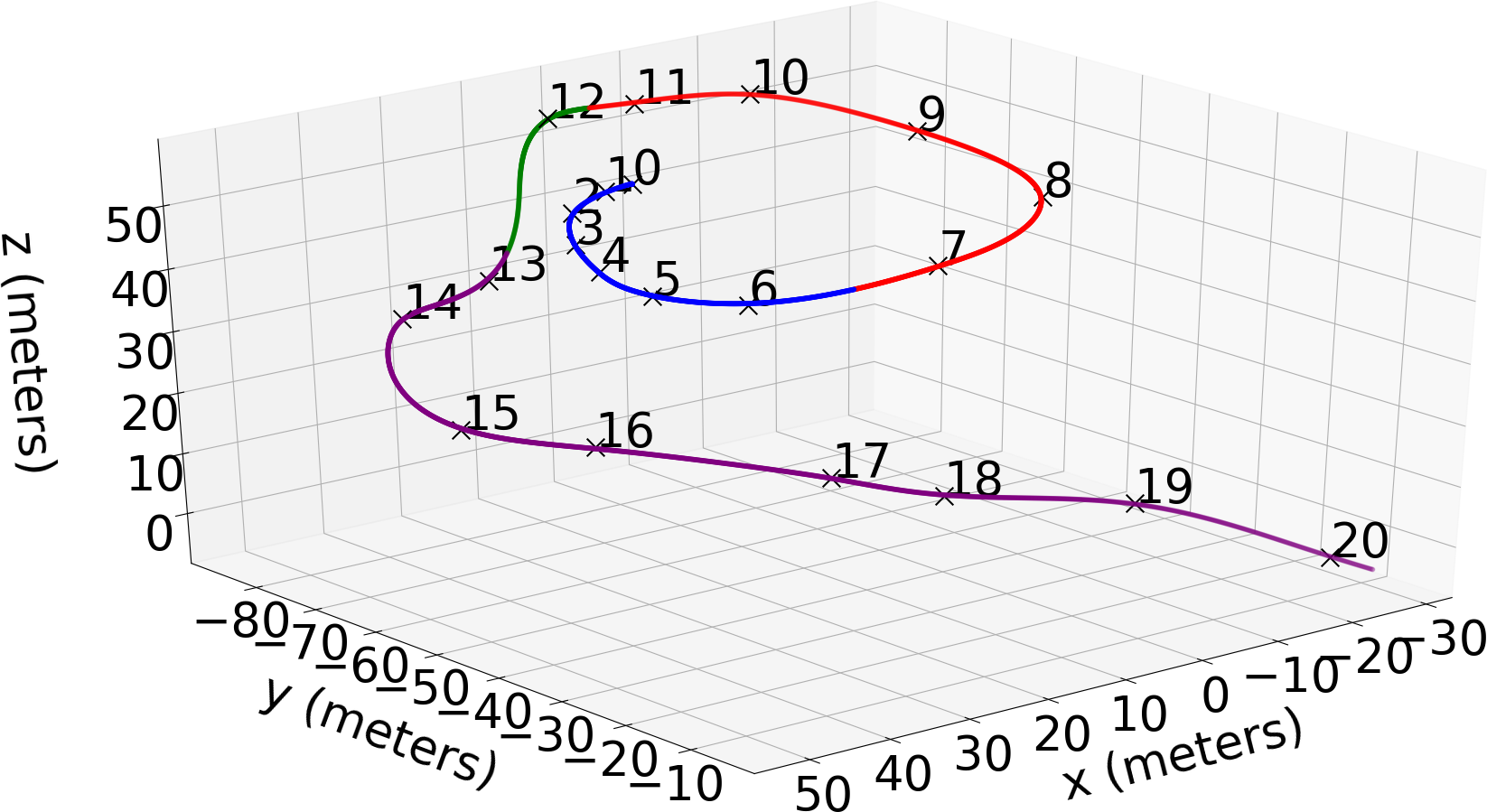}
    \caption{Before starting the sampling, AutoTune segments a trajectory according to the gradient of $z$. The above maneuver was split into flat (blue), ascent (red), drop (green, steep descent), and descent (purple).}
    \label{fig:traj_segmentation}
    \vspace{-3ex}
\end{figure}

Complex high-speed trajectories require different controller behaviors along the track.
For example, consider the reference trajectory illustrated in Fig.~\ref{fig:traj_segmentation}.
The initial segment, depicted in blue, is approximately planar but has a large curvature in the x-y plane.
In contrast, the drop segment, depicted in green in Fig.~\ref{fig:traj_segmentation}, has a large gradient in height, but little motion in the x-y plane.
Clearly, these two segments need different controller behaviors to be successfully tracked.
The blue planar segment requires the controller to be very precise on x-y tracking, but less controller authority is needed on the z-plane.
Conversely, the large drop in altitude of the green segment necessitates very accurate tracking of the reference on the z plane, but less on the x-y one.
Accounting for this behavior is important for time-optimal trajectories, where the drone is always close to its physical limits.
Therefore, global parameters are likely to fail to track the entire trajectory, no matter how many samples are generated.

Motivated by this observation, we split the trajectory into multiple segments according to the height gradient of the reference.
In each segment different parameters are assigned to the controller.
Specifically, we assign each point to the class \emph{flat}, \emph{ascent}, or \emph{descent} if the difference in height $g_z = z[k] - z[k+1]$ with its successor is $|g_z|<1m , g_z \geq 1m, g_z \leq -1m$, respectively.
This segmentation condition is kept fixed for all experiments and ablated in the appendix.
The resulting segments are then clustered such that the minimum segment duration is $2$ seconds.
Eventually, all the descent and ascent segments with a slope higher than \SI{45}{\degree} are recursively split into two equal parts, where the first is assigned to the class \emph{steep} and the second remains assigned to the original class.
Fig.~\ref{fig:traj_segmentation} shows the result of the segmentation algorithm in one of our testing maneuvers.
To account for the strong correlations between segments and keep the optimization global over the trajectory, the controller parameters associated with each segment are updated jointly.
More details about the joint optimization process and other segmentation examples are available in the appendix.

\subsection{Sampler Initialization} \label{subsec:regressors}

The Metropolis-Hastings algorithm requires an initial parameter configuration $w_0$ to initialize the sampling.
Instead of using a random initialization, we propose to use an informed guess for $w_0$.
\rebuttal{Specifically, we use a Gradient Boosting regressor~\cite{friedman2001greedy} with default parameters to predict initial controller parameters for each trajectory segment.}
The training data for this regressor are controller parameters found to be optimal on 5 training trajectories different in layout from the testing ones.
A different regressor is trained for each type of trajectory segment, i.e. flat, ascent, descent, and steep.
Five features including information about the reference trajectory are used for prediction: the number of points in the segment, the slope of the line connecting the first and last point of the segment, as well as their height difference, and the mean velocity and acceleration.
These features have been selected with a cross-validation procedure.

Overall, the idea of predicting an initial guess $w_0$ with a regressor trained on previously seen trajectory experimentally shows to drastically reduce the number of samples (up to 88\%) to find controller parameters for flying a time-optimal trajectory.
More details about the training data, the training procedure, and an ablation study of the features are available in the supplementary material.

\section{Experiments} \label{sec:experiments}

We design our evaluation procedure to address the following questions: Can AutoTune find controller parameters to fly high-speed trajectories? What are the characteristics of the optimization space of controller parameters for the task of high-speed flight? Do the tuned controllers improve performance on a physical platform? Furthermore, we validate our design choices with ablation studies.
We encourage the reader to watch the supplementary video for qualitative results. 

\subsection{Experimental Setup} \label{subsec:experimental_setup}

We use for our experiments two simulators known for their physical and visual realism: Microsoft AirSim \cite{madaan2020airsim}, and Flightmare \cite{yunlong2020flightmare}.
\revision{Our sampling procedure is strongly favored by the high speed at which they can simulate physics (up to 10K times real-time).}
We test AutoTune on six trajectories selected to evaluate controller performance under strong accelerations and high-speed on all axes.
%
%
\begin{table*}[t]
    \centering
    \begin{tabular}{c c c c c c c}
        \toprule
        \multirow{2}{*}{Team} & \multicolumn{3}{c}{Qualification Round} & \multicolumn{3}{c}{Final Round}\\
        \cmidrule(lr){2-4}\cmidrule(lr){5-7}
        & Lap Time [s] & Max Vel [m/s] & Avg Vel [m/s] & Lap Time [s] &
        Max Vel [m/s] & Avg Vel [m/s]\\
        \midrule
        QuetzalC++ & 42.01 & 17.20 & 8.13 & 53.52 & 28.19 & 9.82\\
        Chuchichaschtli & 37.58 & 18.96 & 9.11 & 53.49 & 18.70 & 9.54\\
        Dedale & 30.11 & 16.49 & 11.33 & 39.78 & 20.02 & 12.88\\
        \textbf{AutoTune} & \textbf{24.05} & \textbf{21.68} & \textbf{14.06} & \textbf{38.09} & \textbf{19.83} & \textbf{14.01}\\
        \bottomrule
    \end{tabular}
    \vspace{1em}
    \caption{Game of Drones 2019 leaderboards. AutoTune outperforms the winner of the competition in both qualification and final round.}
    \label{tab:god_leaderboard}
    \vspace{-3ex}
\end{table*}

For comparison, we use \rebuttal{four} baselines for controller tuning.
A naive one (\emph{Random Sampling}) which randomly samples parameters independently and uniformly on all axis with a variance of $5$. 
\rebuttal{Moreover, we compare to the strong baselines of \emph{Bayesian Optimization}~\cite{marco_ICRA_2016} with multiple choices of the Gaussian kernel; \emph{Particle Swarm Optimization}~\cite{kennedy1995pso} (PSO), with $10$ particles and using $0.5, 1, 2$ as, respectively, inertia weight, cognitive constant, and social constant; and \emph{Covariance Matrix Adaptation} Evolution Strategy (CMA-ES)~\cite{hansen2016cma}.
}.
All baselines start tuning from the same point as ours: the trajectory is divided into parts and the regressor predicts initial parameters.
Note that the traditional tuning methods based on gradient-based optimization~\cite{CEDRO2019156, Julka15IES, Astrom83Autom} are impractical for this task, given the difficulty to explicitly find the relationship between the parameters of our receding-horizon MPC controller and the tracking performance over the entire maneuver.

We define the metric of \emph{Trajectory Completion} ($TC$) to compare the different approaches.
Formally, this metric is defined as:
\begin{equation}
    TC = \frac{\sum_{ i \in waypoints} \mathds{1}[i]}{\sum_{ i \in waypoints} 1},
\end{equation}
where the indicator function for waypoint $i$ is 
\begin{equation}
\mathds{1}[i] = 
\begin{cases}
1 &\mbox{if drone at distance $<d$ from $i$ at $t_r(i)$}, \\
0 &\mbox{otherwise,}
\end{cases} \notag
\end{equation}
and $t_r(i)$ is the time when the reference $\boldmath{\tau}_r$  predicts the quadrotor to pass the waypoint $i$.
Whenever a waypoint is missed by more than $d=\SI{1.3}{\meter}$ (gate radius for our drone racing experiments)  or the drone crashes, the experiment is stopped and the metric calculated.
We use this metric in our experiment since it is easy to interpret and can be compared across different experiments.
%
%



\subsection{Tracking Minimum-Time Trajectories} \label{subsec:tracking_minimumtime}

We first evaluate the performance of AutoTune compared to the baselines.
The results are summarized in Table \ref{tab:optimization_results}.
AutoTune is consistently the best across all maneuvers.
For easy maneuvers, e.g. the slow circle or the flip, almost any controller configurations can complete the track, and all methods can find a viable solution.
However, for more difficult maneuvers, the gap between our approach and the baselines widens, reaching up to $90\%$ in the Spiral track.
\rebuttal{On the most difficult tracks, the Bayesian baseline has difficulties fitting the optimization function, while PSO generally gets stuck into a local minimum after few samples.
Given enough samples, CMA-ES achieves very good results, with the exception of the Spiral trajectory, where it consistently fails to pass through the second gate.}
%
%

\begin{figure}[t]
    \centering
    \resizebox{0.73\linewidth}{!}{%
\begin{tikzpicture}

\definecolor{color0}{rgb}{0.8,0.2,0.0666666666666667}

\begin{axis}[
legend cell align={left},
legend style={
  fill opacity=0.8,
  draw opacity=1,
  text opacity=1,
  at={(0.03,0.97)},
  anchor=north west,
  draw=white!80!black
},
tick align=outside,
tick pos=left,
x grid style={white!69.0196078431373!black},
xlabel={Samples},
xmajorgrids,
xmin=-0.2, xmax=26.2,
xtick style={color=black},
y grid style={white!69.0196078431373!black},
ylabel={Trajectory Completion (\%)},
ymajorgrids,
ymin=-11.227763691156, ymax=105.296560175769,
ytick style={color=black}
]
\path [draw=red, fill=red, opacity=0.3]
(axis cs:1,58.75)
--(axis cs:1,3.75)
--(axis cs:2,3.75)
--(axis cs:3,3.75)
--(axis cs:4,3.75)
--(axis cs:5,11.8628903251639)
--(axis cs:6,23.7867965644036)
--(axis cs:7,17.8093626395444)
--(axis cs:8,17.9198719845468)
--(axis cs:9,2.56347586990249)
--(axis cs:10,33.1674917693965)
--(axis cs:11,31.8885933836549)
--(axis cs:12,-5.93120351538663)
--(axis cs:13,47.7108743617003)
--(axis cs:14,53.9628644612183)
--(axis cs:15,21.9686628594442)
--(axis cs:16,56.25)
--(axis cs:17,53.8397459621556)
--(axis cs:18,53.9628644612183)
--(axis cs:19,-4.05137886268713)
--(axis cs:20,17.1688597440489)
--(axis cs:21,54.2264973081037)
--(axis cs:22,55.9175170953614)
--(axis cs:23,53.9628644612183)
--(axis cs:24,47.7108743617003)
--(axis cs:25,47.458471303941)
--(axis cs:25,60.041528696059)
--(axis cs:25,60.041528696059)
--(axis cs:24,64.7891256382997)
--(axis cs:23,63.5371355387817)
--(axis cs:22,64.0824829046386)
--(axis cs:21,65.7735026918963)
--(axis cs:20,67.8311402559511)
--(axis cs:19,51.5513788626871)
--(axis cs:18,63.5371355387817)
--(axis cs:17,71.1602540378444)
--(axis cs:16,61.25)
--(axis cs:15,70.5313371405558)
--(axis cs:14,63.5371355387817)
--(axis cs:13,64.7891256382997)
--(axis cs:12,50.9312035153866)
--(axis cs:11,60.6114066163451)
--(axis cs:10,66.8325082306035)
--(axis cs:9,59.9365241300975)
--(axis cs:8,72.0801280154532)
--(axis cs:7,69.6906373604556)
--(axis cs:6,66.2132034355964)
--(axis cs:5,65.6371096748361)
--(axis cs:4,58.75)
--(axis cs:3,58.75)
--(axis cs:2,58.75)
--(axis cs:1,58.75)
--cycle;

\addplot [line width=2.4pt, blue, opacity=1]
table {%
1 55
2 60
3 60
4 60
5 60
6 60
7 50
8 50
9 60
10 65
11 60
12 60
13 65
14 60
15 65
16 60
17 65
18 65
19 65
20 60
21 100
22 100
23 100
24 100
25 100
};
\addlegendentry{AutoTune}
\addplot [line width=2.4pt, color0, opacity=1]
table {%
1 31.25
2 31.25
3 31.25
4 31.25
5 38.75
6 45
7 43.75
8 45
9 31.25
10 50
11 46.25
12 22.5
13 56.25
14 58.75
15 46.25
16 58.75
17 62.5
18 58.75
19 23.75
20 42.5
21 60
22 60
23 58.75
24 56.25
25 53.75
};
\addlegendentry{Human Experts}
\end{axis}

\end{tikzpicture}%
    }
    \caption{Comparison between AutoTune and a set of human experts on the Qualifier track (Fig.~\ref{fig:traj_segmentation}) in simulation. Despite having access to more information than AutoTune, no human was able to find suitable parameters.}
    \label{fig:autotune_vs_expert}
    \vspace{-3ex}
\end{figure}
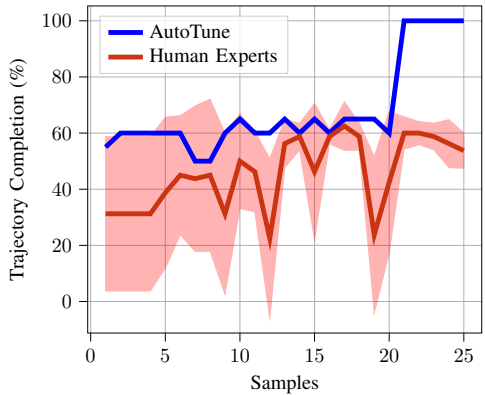
Fig.~\ref{fig:autotune_vs_expert} shows the comparison between the tuning performance of AutoTune and four human experts over time.
All humans are graduate students expert in quadrotor research and not authors of this paper.
They have additional sources of information with respect to our algorithm, i.e. the 3D trajectory flown by the drone.
The results show that tuning parameters by hand is extremely challenging for humans, given the non-intuitive relationship between parameters and performance.
%
%


We additionally compare our approach to the top three methods in the 2019 AirSim Game of Drones competition~\cite{madaan2020airsim}.
We compare the methods both on the qualification and final round of the competition.
The results of this experiment are summarized in Table~\ref{tab:god_leaderboard}.
On the qualifier track flying AutoTune achieves a lap-time of $\SI{24.05}{\second}$, while the winner only reaches the goal in $\SI{30.11}{\second}$, with a lap-time $25\%$ longer than ours.
Also in the final round, AutoTune outperforms the winner of the competition with a $\SI{1.7}{\second}$ margin, completing the track in approximately $5\%$ less time.
Interestingly, our approach converges to a policy with a maximum speed not necessarily higher than others.
However, we achieve an average velocity higher than baselines, and therefore a faster lap-time. 
This experiment shows that tuning controller parameters with an automated procedure allow quadrotors to fly faster trajectories.
\subsection{Application in the Real World}
\begin{figure*}[t]
    \centering
    \begin{tabular}{ccc}
    \resizebox{0.28\linewidth}{!}{
\begin{tikzpicture}

\definecolor{color0}{rgb}{1,0.498039215686275,0.0549019607843137}
\definecolor{color1}{rgb}{0.635,0.078,0.184}
\definecolor{color2}{rgb}{0,0.447,0.741}

\begin{axis}[
tick align=outside,
tick pos=left,
title={Mean Tracking Error [m]
},
x grid style={white!69.0196078431373!black},
xmin=0.5, xmax=2.5,
xtick style={color=black},
xtick={1,2},
xticklabels={Default,AutoTune},
y grid style={white!69.0196078431373!black},
ymajorgrids,
ymin=0.6555, ymax=0.7545,
ytick style={color=black}
]
\addplot [line width=1pt, black]
table {%
1 0.715
1 0.7
};
\addplot [line width=1pt, black]
table {%
1 0.74
1 0.75
};
\addplot [line width=1pt, black]
table {%
0.875 0.7
1.125 0.7
};
\addplot [line width=1pt, black]
table {%
0.875 0.75
1.125 0.75
};
\addplot [line width=1pt, black]
table {%
2 0.675
2 0.66
};
\addplot [line width=1pt, black]
table {%
2 0.705
2 0.72
};
\addplot [line width=1pt, black]
table {%
1.875 0.66
2.125 0.66
};
\addplot [line width=1pt, black]
table {%
1.875 0.72
2.125 0.72
};
\path [draw=black, fill=color1, line width=1pt]
(axis cs:0.75,0.715)
--(axis cs:1.25,0.715)
--(axis cs:1.25,0.74)
--(axis cs:0.75,0.74)
--(axis cs:0.75,0.715)
--cycle;
\path [draw=black, fill=color2, line width=1pt]
(axis cs:1.75,0.675)
--(axis cs:2.25,0.675)
--(axis cs:2.25,0.705)
--(axis cs:1.75,0.705)
--(axis cs:1.75,0.675)
--cycle;
\addplot [line width=1pt, color0]
table {%
0.75 0.73
1.25 0.73
};
\addplot [line width=1pt, color0]
table {%
1.75 0.69
2.25 0.69
};
\end{axis}

\end{tikzpicture}} &
    \resizebox{0.28\linewidth}{!}{
\begin{tikzpicture}

\definecolor{color0}{rgb}{1,0.498039215686275,0.0549019607843137}
\definecolor{color1}{rgb}{0.635,0.078,0.184}
\definecolor{color2}{rgb}{0,0.447,0.741}

\begin{axis}[
tick align=outside,
tick pos=left,
title={Max Tracking Error [m]
},
x grid style={white!69.0196078431373!black},
xmin=0.5, xmax=2.5,
xtick style={color=black},
xtick={1,2},
xticklabels={Default,AutoTune},
y grid style={white!69.0196078431373!black},
ymajorgrids,
ymin=1.21835, ymax=1.47465,
ytick style={color=black}
]
\addplot [line width=1pt, black]
table {%
1 1.404
1 1.396
};
\addplot [line width=1pt, black]
table {%
1 1.4375
1 1.463
};
\addplot [line width=1pt, black]
table {%
0.875 1.396
1.125 1.396
};
\addplot [line width=1pt, black]
table {%
0.875 1.463
1.125 1.463
};
\addplot [line width=1pt, black]
table {%
2 1.25
2 1.23
};
\addplot [line width=1pt, black]
table {%
2 1.275
2 1.28
};
\addplot [line width=1pt, black]
table {%
1.875 1.23
2.125 1.23
};
\addplot [line width=1pt, black]
table {%
1.875 1.28
2.125 1.28
};
\path [draw=black, fill=color1, line width=1pt]
(axis cs:0.75,1.404)
--(axis cs:1.25,1.404)
--(axis cs:1.25,1.4375)
--(axis cs:0.75,1.4375)
--(axis cs:0.75,1.404)
--cycle;
\path [draw=black, fill=color2, line width=1pt]
(axis cs:1.75,1.25)
--(axis cs:2.25,1.25)
--(axis cs:2.25,1.275)
--(axis cs:1.75,1.275)
--(axis cs:1.75,1.25)
--cycle;
\addplot [line width=1pt, color0]
table {%
0.75 1.412
1.25 1.412
};
\addplot [line width=1pt, color0]
table {%
1.75 1.27
2.25 1.27
};
\end{axis}

\end{tikzpicture}} &
    \resizebox{0.28\linewidth}{!}{\input{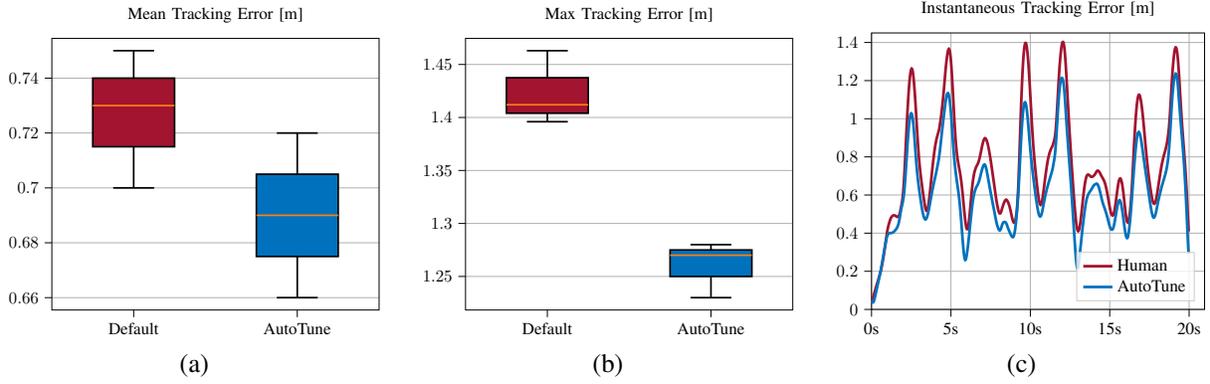}} \\
       (a) &
       (b) &
       (c) \\
    \end{tabular}
    \caption{Results in the real world. After tuning the parameters in simulation, we evaluate the best configuration found by AutoTune on a physical platform. We compare the performance with the parameters tuned by a human. We perform three runs for each parameter set. We also report the error as a function of time (c) for the best run of each approach.}
    \label{fig:auto_rw}
\end{figure*}
AutoTune can be used to tune the controller of a physical platform.
To do so, we compute a minimum-time trajectory double Split-S trajectory of 21 waypoints~\cite{Foehn2020}. This trajectory is used to tune the controller in the Flightmare simulator.
\revision{The resulting controller is then evaluated on a physical platform in a tracking arena of volume $30\times30\times8\SI{}{\meter}$, where the quadrotor achieves speeds over $\SI{50}{\kilo\meter\per\hour}$}.
Figure~\ref{fig:auto_rw} shows the results of this experiment.
AutoTune improves average tracking error by $6\%$ and decreases the maximum displacement from the reference by $12\%$. 
In addition, our controller parameters give more consistent performance over multiple runs than the baseline.

\subsection{Robustness to Changes in Mass, Velocity, and Track Layout} \label{subsec:robustness2changes}
\begin{figure*}
    \centering
    \begin{tabular}{ccc}
    \setlength{\tabcolsep}{0pt}
     \includegraphics[width=0.3\linewidth]{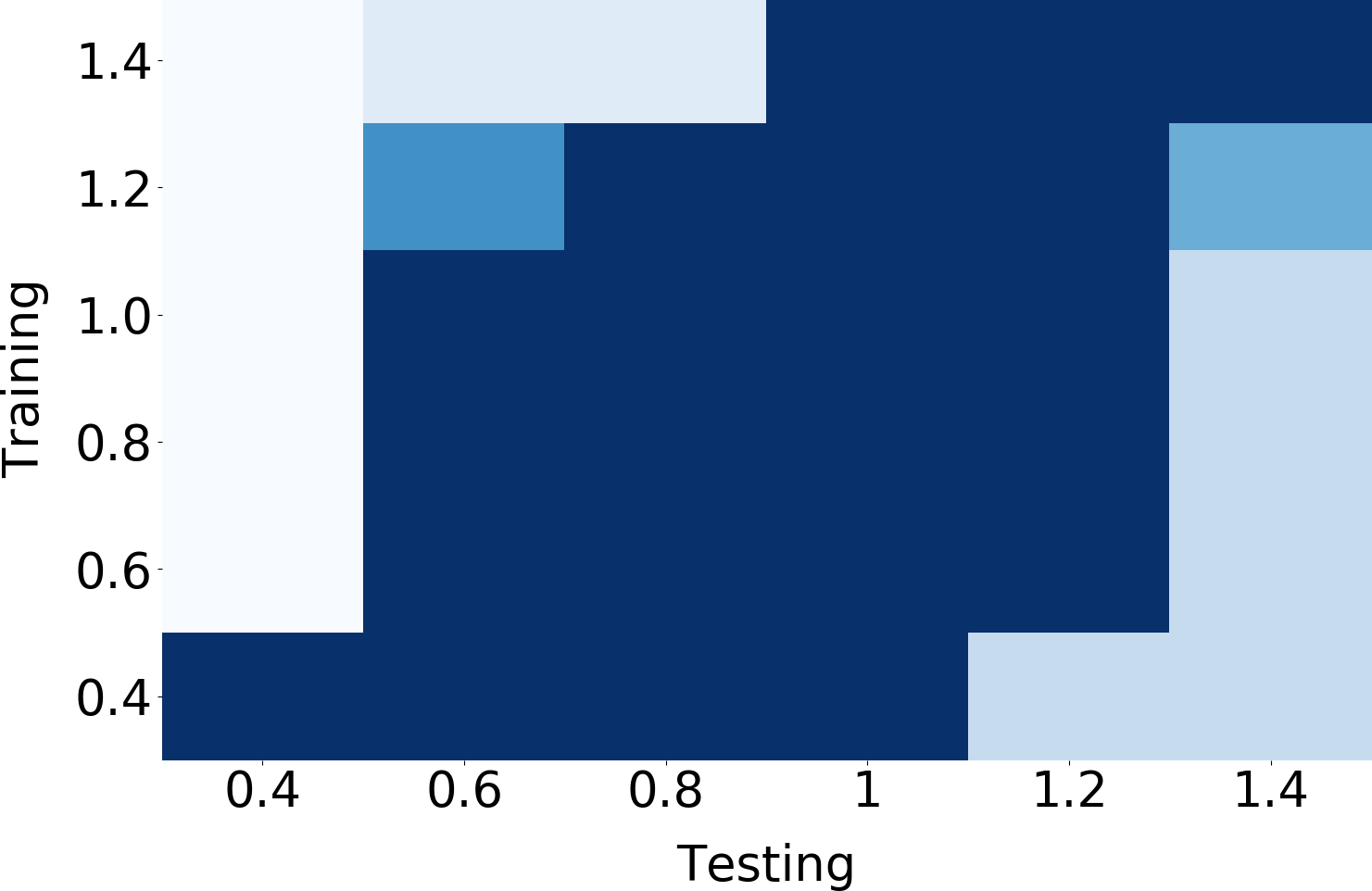} &
     \includegraphics[width=0.3\linewidth]{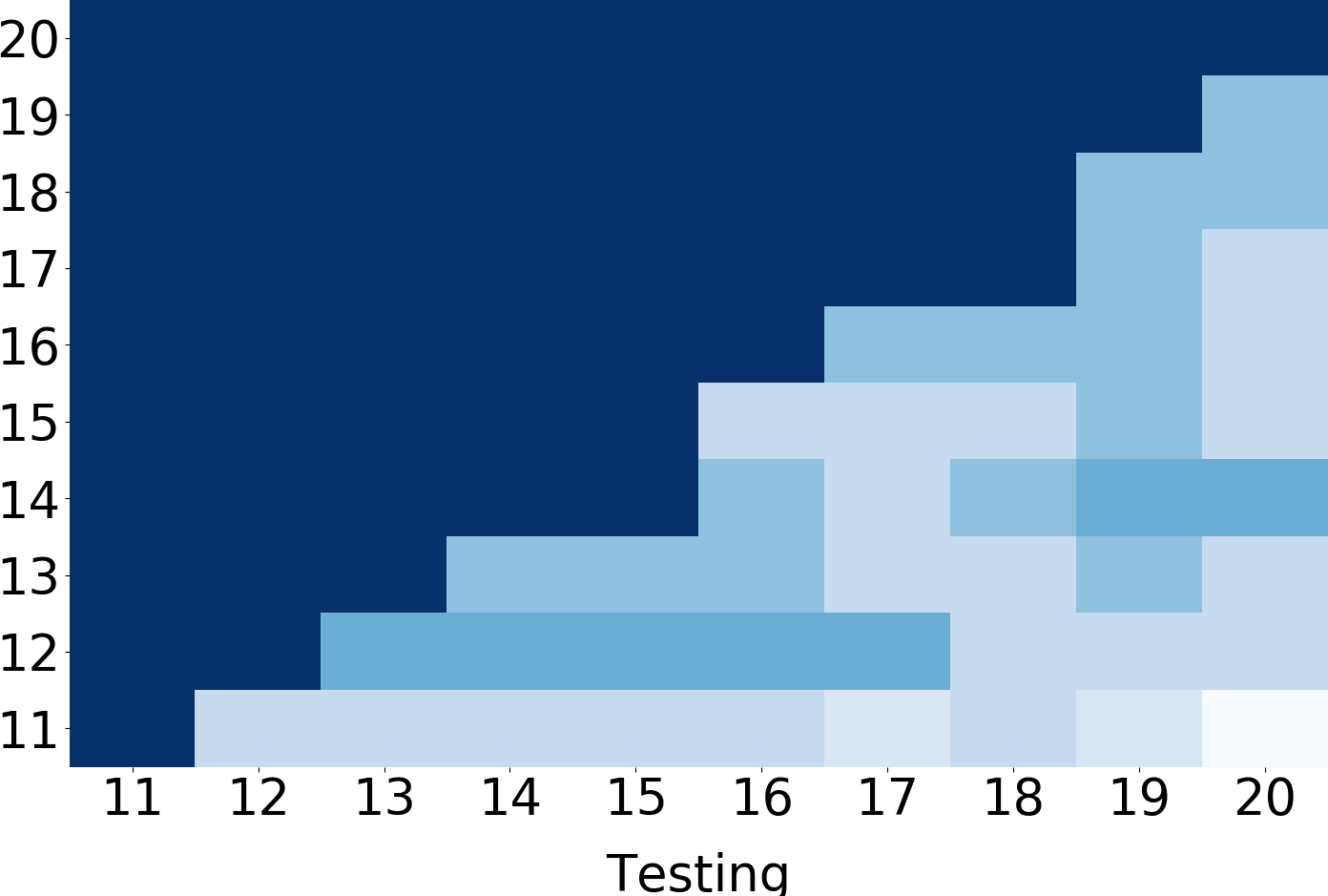} & 
     \includegraphics[width=0.34\linewidth]{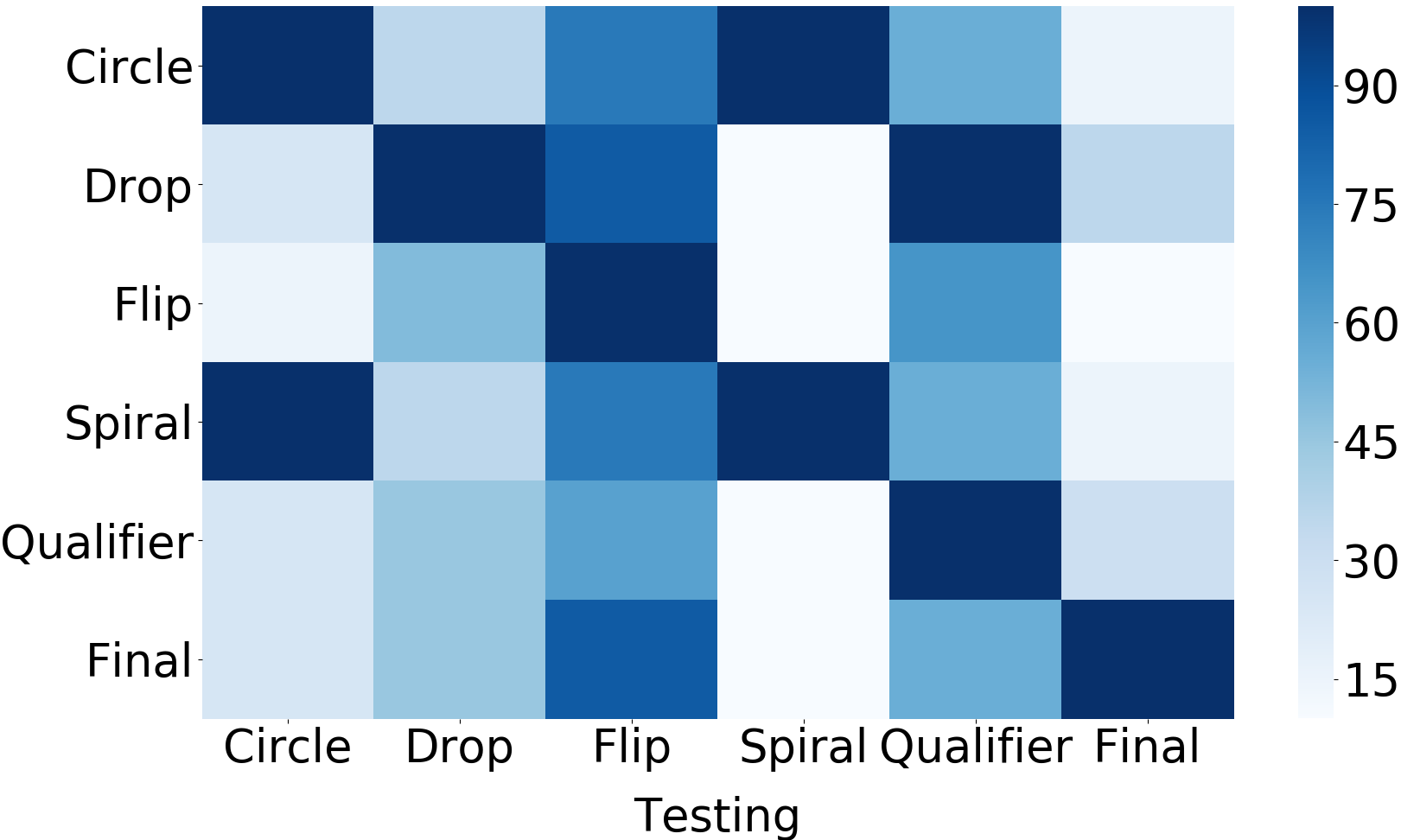} \\
     (a) Quadrotor Mass [$\SI{}{\kilo\gram}$] & (b) Max velocity [$\SI{}{\meter\per\second}$] & (c) Track Layout \\
    \end{tabular}
    \caption{Performance analysis when parameters used for tuning (Training) is different to the one used during execution (Testing). Overall, the parameters are robust to imperfect identification of the mass, and the faster the maneuver, the more sensitive the controller is to the parameters. However, the controllers require to be tuned specifically to the maneuver. When the training and testing maneuvers are different, performance generally drops. }
    \label{fig:robustness_to_changes}
\end{figure*}
\begin{figure*}[h]
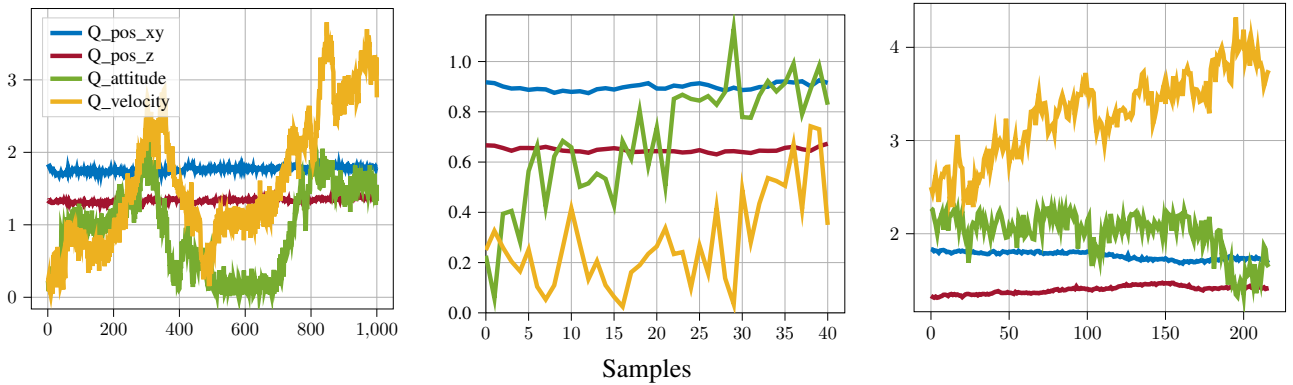

\centering
\begin{tabular}{ccc}
    \resizebox{0.3\linewidth}{!}{\input{images/tickzfiles/segment_5}} &
    \resizebox{0.3\linewidth}{!}{
\begin{tikzpicture}

\definecolor{color0}{rgb}{0,0.447,0.741}
\definecolor{color1}{rgb}{0.635,0.078,0.184}
\definecolor{color2}{rgb}{0.466,0.674,0.188}
\definecolor{color3}{rgb}{0.929,0.694,0.125}

\begin{axis}[
tick align=outside,
tick pos=left,
x grid style={white!69.0196078431373!black},
xmajorgrids,
xmin=0, xmax=42,
xtick style={color=black},
xtick={0,5,10,15,20,25,30,35,40},
y grid style={white!69.0196078431373!black},
ymajorgrids,
ymin=0, ymax=1.18565115391914,
ytick style={color=black},
ytick={-0.2,0,0.2,0.4,0.6,0.8,1,1.2},
yticklabels={−0.2,0.0,0.2,0.4,0.6,0.8,1.0,1.2}
]
\addplot [line width=2.4pt, color0, opacity=1]
table {%
0 0.917431192660551
1 0.914055288966916
2 0.900765168167035
3 0.892618442572325
4 0.894027970966851
5 0.887455383969714
6 0.890927551334292
7 0.889218976895281
8 0.875593348344389
9 0.883752558395345
10 0.879283981136758
11 0.881782726692639
12 0.874398380857895
13 0.889810353125238
14 0.893885156413503
15 0.888564587339504
16 0.897097517532585
17 0.902620555442811
18 0.906473649484284
19 0.913829320852925
20 0.892582516782349
21 0.892088327683906
22 0.904263200236319
23 0.900101317042459
24 0.909857194954728
25 0.913472193911215
26 0.906347241735602
27 0.894268679254028
28 0.884329712332956
29 0.896003437476083
30 0.886558872824561
31 0.888909059258251
32 0.898335096753951
33 0.899660344430783
34 0.919463874502826
35 0.920512203195963
36 0.916596119152029
37 0.92155286811833
38 0.901700292053791
39 0.92542151310547
40 0.915115645607828
};
\addplot [line width=2.4pt, color1, opacity=1]
table {%
0 0.666666666666667
1 0.665149891475292
2 0.65596138314782
3 0.645073835988256
4 0.655933019432627
5 0.655985119426059
6 0.654708378963795
7 0.660525747949459
8 0.652413624377241
9 0.645440295693241
10 0.642845072383716
11 0.642455916691631
12 0.636937957586709
13 0.648736844188648
14 0.651859519516329
15 0.655169547990929
16 0.650552922366569
17 0.639823853321088
18 0.640885128307416
19 0.643040873073224
20 0.643296728006379
21 0.643291935922272
22 0.642894056846576
23 0.638158081278833
24 0.640339035559761
25 0.647476451053859
26 0.637195650760759
27 0.630594022765336
28 0.642473974378691
29 0.643387645247961
30 0.639678544408305
31 0.636154078817623
32 0.645480143332227
33 0.644555750644175
34 0.645093410506285
35 0.656555485606781
36 0.660549042373452
37 0.651156980633571
38 0.645177364164805
39 0.662798075277379
40 0.673039698488231
};
\addplot [line width=2.4pt, color2, opacity=1]
table {%
0 0.227272727272727
1 0.0587948367796646
2 0.39395382126502
3 0.405358472248739
4 0.283672735285777
5 0.563988802862111
6 0.663975702346414
7 0.421951893363505
8 0.622224936143182
9 0.683893885185357
10 0.658934788104525
11 0.502098135686448
12 0.515155171482725
13 0.553795592394307
14 0.533434872274159
15 0.427331674871336
16 0.682414390032434
17 0.613615745619766
18 0.803013762447155
19 0.584805685234973
20 0.737013124216791
21 0.50427309385155
22 0.85160105299647
23 0.867883467638395
24 0.851216597311252
25 0.844519456318984
26 0.862287819213261
27 0.828313372605132
28 0.879487356341625
29 1.13038275987183
30 0.778983633446805
31 0.776245016039241
32 0.864835712286311
33 0.921524618554052
34 0.881268840457127
35 0.913948493032648
36 0.991239274713814
37 0.791018555714589
38 0.890251359400361
39 0.982527862214093
40 0.828311198748368
};
\addplot [line width=2.4pt, color3, opacity=1]
table {%
0 0.25
1 0.326031610406277
2 0.25984376382718
3 0.204368017705032
4 0.165902728902032
5 0.251741200335775
6 0.10430638776555
7 0.0532487390098192
8 0.109620651941207
9 0.260165480447258
10 0.414965407813027
11 0.272470684241617
12 0.134008841035399
13 0.205813352160337
14 0.110129954067445
15 0.0637494562302563
16 0.0250148789256385
17 0.161462614046384
18 0.18958357873846
19 0.234701019410707
20 0.264461241255952
21 0.336587762824325
22 0.234078115833619
23 0.241556192346007
24 0.100707630129274
25 0.267986985330887
26 0.158203349143292
27 0.417654610078212
28 0.137386830208449
29 0.0348960325594295
30 0.489878852778303
31 0.28144130497206
32 0.434707904091527
33 0.536435441006337
34 0.524704999509885
35 0.505873319006165
36 0.670075659675398
37 0.413523186163935
38 0.74216074617262
39 0.730746593960607
40 0.349977689581185
};
\end{axis}

\end{tikzpicture}} &
    \resizebox{0.3\linewidth}{!}{\input{images/tickzfiles/segment_6}} \\
    \multicolumn{3}{c}{Samples}
\end{tabular}
    \caption{\rebuttal{Robustness to different initial conditions. The parameters' values are divided by their maximum value achieved over all runs. AutoTune converges to a configuration with 100\% trajectory completion for all initial conditions}. Initialization strongly affects the converge speed. All runs converge to a different optimum, demonstrating the multi-modal characteristic of the optimization function.}
    \vspace{-3ex}
    \label{fig:exp_hairpin}
\end{figure*}

In this section, we study the robustness of the parameters found by AutoTune to changes in drone's mass, flight speed, and track layout.
All experiments are made on the Qualifier track.
Figure~\ref{fig:robustness_to_changes} show the results of these experiments.
The parameters are overall robust to changes in the quadrotor mass and can complete the task even for very different settings.
%
%
When changing the maximum speed achieved during flight (Fig.~\ref{fig:robustness_to_changes}-b), we observe that a faster trajectory requires more precise tuning.
%
%
Finally, we test whether parameters generalize between different maneuvers.
To favor generalization, we copy parameters for each segment independently.
The results (Fig.~\ref{fig:robustness_to_changes}-c) show that, when the maneuver used for tuning is different from the testing one, the performance generally drops.
\rebuttal{
One possible solution to this problem would be to do automatic fine-grained segmentation of trajectories. 
Indeed, we hypothesize that smaller trajectory segments, despite being more difficult to optimize, would transfer better between different layouts.
}

\subsection{Robustness to Initial Conditions} \label{subsec:initializations}
In this section, we study the evolution of the controller parameters during optimization for different initialization conditions.
Specifically, we start the sampling procedure from 3 random initializations and tune the controller with our approach in the Flightmare simulator on the \emph{Qualifier} track (Fig.~\ref{fig:traj_segmentation}).
Figure~\ref{fig:exp_hairpin} shows the results of this experiment.
\rebuttal{AutoTune finds parameters to reach 100\% of trajectory completion for all initial conditions.
However, while some require as little as $50$ samples, others require up to $1$K to converge.
Interestingly, the approach follows different paths in the optimization space for every initialization.
In addition, the sampling converges to a different local optimum for each initial condition.
This behavior empirically shows that the relationship between parameters and performance at high-speed is \emph{multi-modal}, \emph{i.e.} different controller parameters have the same performance.}
These characteristics of the optimization function represent a challenge for gradient-based and Bayesian methods~\cite{Basri15TIMC, berkenkamp2016safe, berkenkamp2016safe}, which tend to converge to the mean between different optima.
Conversely, since Metropolis-Hastings sampling can approximate any probability distribution under relatively mild assumptions, our approach does not suffer from the multi-modal nature of the optimization function.

\subsection{Ablation Studies} \label{subsec:ablation}
\begin{table}[t]
    \centering
    \begin{tabular}{l  c c}
        \toprule
          & Trajectory Completion [\%] & samples \\
        \midrule
        AutoTune & \textbf{100} & \textbf{21} \\
        -- Regressor & 100 & 172 \\
        -- Segmentation & 65 & 300 \\
        \bottomrule
    \end{tabular}
    \caption{Ablation Study of the system's component on the Qualifier trajectory.}
    \label{tab:blocks_ablationstudy}
    \vspace{-3ex}
\end{table}

AutoTune is based on several components to reduce the sample complexity of Metropolis-Hastings sampling.
We now validate our design with an ablation study.
In particular, we ablate the following components: (i) the use of a regressor for predicting an initial controller to initialize sampling, (ii) the segmentation of the trajectory in different parts.
The results in Table~\ref{tab:blocks_ablationstudy} show that all components are important, but some have a larger impact than others.
The initial guess produced by the regressor drastically reduces the number of samples to convergence, making the sampler find a solution in $88\%$ less time.
However, the most important contribution comes from the trajectory segmentation.
Without this component, the sampler cannot find parameters to complete more than $65\%$ of the trajectory in less than $300$ samples.
This is because global parameters do not allow the controller to dynamically adapt to different parts of the trajectory.

\section{Discussion and Conclusions} \label{cha:conclusions}

This paper shows the importance of an automated tuning procedure to fly high-speed maneuvers.
While the effect of tuning is less prominent at low speeds, it acquires a fundamental role when the quadrotor flies at the limits of handling.
In such cases, the relation between the parameters and flight performance (measured, for example, in terms of trajectory completion or tracking error) is non-convex, not injective, and multi-modal.
%
%
In this paper, we propose a sampling-based approach specifically tailored to the task of high-speed flight.
%

One limitation of the proposed approach is that it does not consider closed-loop stability during optimization.
\revision{While prior work proposed a series of techniques to guarantee stability during tuning~\cite{berkenkamp2016safe, marco_ICRA_2016, Basri15TIMC}, such techniques either require a very accurate model of the platform or very conservative parameters exploration strategies.
These makes them suited for tasks like hovering or low-speed flight but not to high-speed flight, where model mismatch makes the parameter's optimization landscape very complex.}
Similar to previous work on agile flight~\cite{kaufmann2020RSS}, we have addressed this problem by tuning the controller exclusively in simulation and directly using the tuned controller on a physical platform.
However, such a strategy strongly depends on the quality of the simulation environment.
Therefore, combining existing techniques for safe tuning with our approach, to either tune from scratch or only finetune the controller on the physical platform, is a very exciting venue for future work.
%
%

{\small
\bibliographystyle{IEEEtran}
\bibliography{references}
}


\end{document}